\documentclass[journal,twoside,web]{ieeecolor}
\usepackage{tmi}
\usepackage{cite}
\usepackage{amsmath,amssymb,amsfonts}
\usepackage{algorithmic}
\usepackage{graphicx}
\usepackage{textcomp}

\usepackage{adjustbox}
\usepackage{booktabs}
\usepackage{multirow}
\usepackage{makecell}
\usepackage{hyperref}
\hypersetup{
    colorlinks=true,
    linkcolor=blue,
    urlcolor=blue,
    citecolor=blue
}

\def\BibTeX{{\rm B\kern-.05em{\sc i\kern-.025em b}\kern-.08em
    T\kern-.1667em\lower.7ex\hbox{E}\kern-.125emX}}
\begin{document}
\title{Semi-Supervised Virtual Staining via Morphology Preservation and Histopathological Realism Constraints}
\author{Baoshun Wang, Weiping Lin, Linwu Wang, Yihuang Hu, Baptiste Magnier, Liansheng Wang, \IEEEmembership{Member, IEEE}
\thanks{This work was supported by the National Natural Science Foundation of China under Grant 62371409, Fujian Provincial Natural Science Foundation of China under Grant 2023J01005. (Corresponding author: Liansheng Wang.)}
\thanks{Baoshun Wang and Weiping Lin contributed equally.}
\thanks{Baoshun Wang, Weiping Lin, Linwu Wang, Yihuang Hu and Liansheng Wang are with Department of Computer Science, School of Informatics, Xiamen University, Xiamen 361005, China (e-mail: \{bswang, wplin, lwwang, huyihuang\}@stu.xmu.edu.cn; lswang@xmu.edu.cn).}
\thanks{Baptiste Magnier is with EuroMov Digital Health in Motion, Univ Montpellier, IMT Mines Ales, Ales, France, and also with Service de Médecine Nucléaire, Centre Hospitalier Universitaire de Nîmes, Université de Montpellier, Nîmes, France (e-mail: baptiste.magnier@mines-ales.fr).}
}

\maketitle

\begin{abstract}
Virtual staining aims to computationally generate target-stained
histopathological images while reducing the cost and time associated
with conventional staining procedures. 
However, existing methods rely
predominantly on strictly paired and accurately registered training
data, which are difficult and expensive to obtain in routine practice.
To reduce this dependence, we propose a stable semi-supervised virtual
staining framework that jointly exploits both limited paired data and
abundant unpaired source images. Directly incorporating unpaired images
is challenging because their generated results lack corresponding
targets for supervision, potentially leading to unrealistic staining,
morphological degradation, or even training collapse. To obtain
reliable supervision from these images, Hessian-derived morphology
preservation extracts structural cues from each source image and
constrains the generated output to retain tissue morphology.
Histopathological realism constraints further guide the output toward plausible target-stain characteristics, preventing the source-derived structural supervision from degenerating into contour enhancement or simple color transformation. Together, the two components suppress
structural and appearance drift, stabilize semi-supervised stain
translation, and promote the preservation of diagnostically relevant information. 
Extensive experiments on H\&E-to-IHC translation for
Ki67 and HER2, as well as FFPE-to-H\&E translation, demonstrate
consistent improvements in image quality, morphology preservation,
robustness, and downstream diagnostic performance. Code will be
available.
\end{abstract}


\section{Introduction}
\label{sec:introduction}

\IEEEPARstart{H}{istopathological} staining reveals tissue characteristics
that are essential for disease diagnosis and treatment planning.
Hematoxylin and eosin (H\&E) staining primarily presents tissue
organization, cellular morphology, and nuclear appearance
\cite{titford2005long}, whereas immunohistochemistry (IHC) visualizes
the expression and spatial distribution of specific biomarkers, such as
HER2 and Ki67 \cite{ramos2014tissue}. These complementary staining
patterns provide important evidence for tumor characterization and
therapeutic decision-making. Conventional staining, however, involves
multiple tissue-processing and chemical procedures, resulting in
substantial labor, reagent consumption, and turnaround time. 
These limitations have motivated computational methods that generate a desired staining modality directly from an available pathological image,  including conditional generative adversarial networks
\cite{rivenson2019virtual,bai2023deep}.

\begin{figure}[t]
  \centering
  \includegraphics[width=1.0\linewidth]{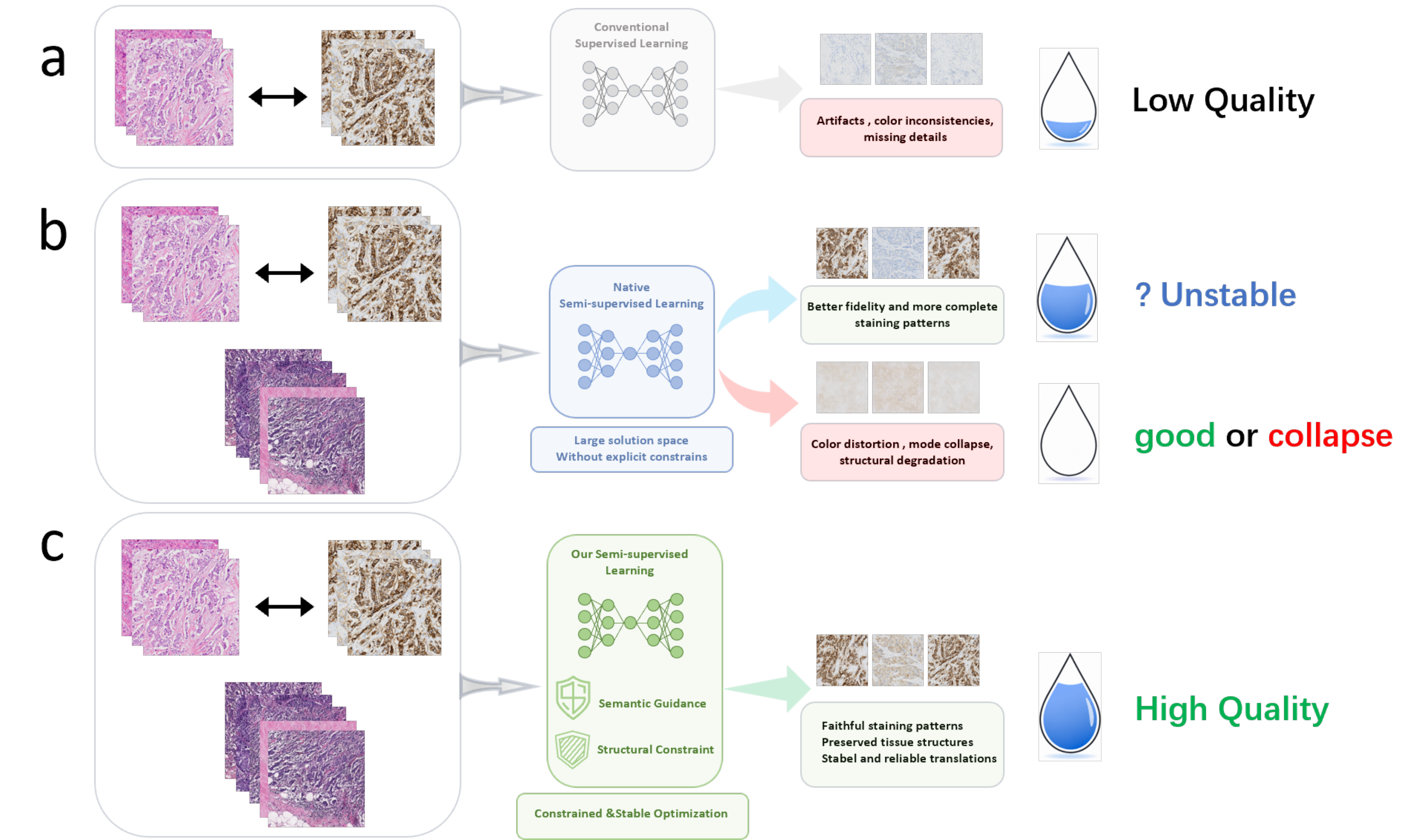}
  \caption{
  Motivation of the proposed framework.
  (a) Limited paired data constrain virtual staining performance.
  (b) Unpaired source images improve data utilization but lack direct
  cross-stain supervision.
  (c) Hessian-derived morphology preservation extracts structural
  supervision from source images, while histopathological realism
  constraints promote realistic target staining.
  }
  \label{fig:motivation}
\end{figure}

With the digitization of pathological slides, virtual staining has
emerged as a computational approach for translating tissue images
between staining modalities \cite{lin2025virtual}. 
Unpaired image-to-image translation methods, such as CycleGAN, CUT, and
pathology-specific stain-transfer models, learn from independently
collected source- and target-domain images without exact spatial
correspondence
\cite{zhu2017unpaired,park2020contrastive,liu2021unpaired}.
This flexibility is attractive in pathology, where corresponding tissue
sections are often unavailable. Nevertheless, virtual staining is not
simply a style-transfer problem. A visually realistic image is
insufficient if tissue boundaries are distorted or biomarker-related
staining appears in inappropriate regions. Because distribution-level
matching cannot determine the correct local relationship between source
tissue and target staining, paired supervision remains important for
preserving morphology and diagnostically relevant staining patterns.
Accordingly, supervised models such as Pix2Pix and Pix2PixHD remain
widely used foundations for virtual staining
\cite{isola2017image,wang2018high}.

The reliability of paired supervision comes with a substantial
data-acquisition burden. Cross-stain image pairs are commonly obtained
from adjacent tissue sections or from tissue imaged before and after
staining. Tissue deformation, missing regions, sectioning artifacts,
staining variations, and registration errors can disrupt their spatial
correspondence. Constructing a high-quality paired dataset therefore
requires extensive registration, meticulous manual inspection, and exclusion of
poorly aligned samples. Consequently, only a limited proportion of
collected images can be retained as registered source--target pairs,
whereas many independently acquired source-domain images remain
available without corresponding target stains.

Semi-supervised learning has achieved considerable success in medical
image classification, segmentation, and related tasks by combining
limited annotated data with additional unannotated samples
\cite{cheplygina2019not,zhou2023semi,berthelot2019mixmatch,xie2020unsupervised}.
This paradigm is particularly promising for virtual staining because
paired images are difficult to acquire, while unpaired source images are
abundant. In this work, \emph{unpaired images} refer specifically to
source-domain images without corresponding target-stain images. Paired
samples provide direct cross-stain supervision, including tissue
correspondence and target-specific staining information, whereas
unpaired samples increase the diversity of tissue appearances available
during training.

Exploiting unpaired images in a generative task is nevertheless
challenging. 
Unlike classification or segmentation, virtual staining must predict a high-dimensional target image, and an unpaired input provides no corresponding target with which to verify the generated result. 
Without appropriate supervision, the generator may deviate from
the intended stain mapping, distort source-tissue morphology, produce
unrealistic target staining, or even collapse during training. 
Consequently, reliable supervision must therefore be derived for unpaired samples without requiring additional target images.

To this end, we first introduce Hessian-derived morphology preservation
to extract structural supervision directly from each source image.
Second-order spatial responses capture morphology-related transitions
around tissue interfaces, cellular boundaries, and nuclear contours and
can be compared across staining modalities. This source-derived signal
therefore constrains tissue morphology even when a paired target is
unavailable. However, morphology preservation alone may encourage the
generator to overemphasize contours or reduce stain translation to a
simple color transformation, without ensuring a realistic target-stain
appearance.

We consequently introduce histopathological realism constraints (HRC)
as a complementary source of supervision. Based on a frozen CONCH
vision (a language model \cite{lu2024conch}), HRC guides generated images
toward plausible target-stain characteristics. For paired samples, the
corresponding target image provides additional cross-stain
representation guidance; for unpaired samples, the target-stain prompt
serves as a stable reference. HRC therefore complements the
source-derived morphology supervision and discourages structure-only
shortcuts. Together, HRC and morphology preservation regulate both
target-stain realism and source-tissue morphology, thereby stabilizing
semi-supervised optimization and promoting the preservation of
diagnostically relevant information.

The proposed framework uses a shared generator for paired and unpaired
samples. Paired samples are optimized using conventional cross-stain
supervision together with HRC and morphology preservation, whereas
unpaired samples obtain supervision from the latter two components. We
evaluate the framework on H\&E-to-IHC virtual staining for Ki67 and
HER2, as well as FFPE-to-H\&E translation, under different
unpaired-to-paired data ratios. The evaluation covers image quality,
morphology preservation, pathology-informed measurements, robustness,
and downstream diagnostic performance.

The main contributions of this work are summarized as:
\begin{itemize}

    \item We propose a stable semi-supervised virtual staining framework
    that jointly exploits limited registered source--target pairs and
    abundant unpaired source images, addressing optimization instability
    caused by the absence of corresponding targets.

    \item We introduce Hessian-derived morphology preservation, which
    obtains structural supervision directly from source images and
    reduces alterations to tissue interfaces, cellular boundaries, and
    nuclear contours during stain translation.

    \item We develop histopathological realism constraints based on a
    frozen CONCH model to promote plausible target-stain characteristics
    and prevent source-derived structural supervision from degenerating
    into contour enhancement or simple color transformation.

    \item We validate the proposed framework on three virtual staining
    tasks under different unpaired-to-paired data ratios, demonstrating
    improvements in staining quality, morphology preservation,
    robustness, and downstream diagnostic performance.

\end{itemize}

\section{Related Work}

\subsection{Virtual Staining}

Virtual staining computationally translates pathological images from
one staining modality to another. Existing approaches can generally be
categorized according to whether spatially corresponding cross-stain
images are available during training.

Unpaired methods learn from independently collected both source- and target-domain images without requiring spatial registration. 
CycleGAN
uses cycle consistency to encourage content preservation, whereas CUT
maintains correspondence between input and output representations
through contrastive learning
\cite{zhu2017unpaired,park2020contrastive}.
Although these approaches reduce the need for paired data, their
supervision is mainly derived from domain-level distribution matching
or representation consistency. They may therefore reproduce a realistic
target-domain appearance without guaranteeing correct biomarker
localization or faithful morphology preservation.

Paired methods use registered source--target images to directly
supervise the desired stain mapping. Pix2Pix and Pix2PixHD combine
reconstruction and adversarial objectives and remain common supervised
backbones
\cite{isola2017image,wang2018high}.
Subsequent studies have explored multi-scale and high-resolution
generation
\cite{liu2022bci,sun2023bi,ma2023efficient};
adaptive, weakly supervised, or weakly paired learning
\cite{li2023adaptive,li2024virtual,guan2025supervised};
pathology-informed objectives
\cite{peng2024advancing,hu2024boosting};
diffusion-based stain transfer
\cite{he2024pst,grosskopf2025histdist};
virtual multiplexing
\cite{pati2024accelerating};
and foundation model-guided translation
\cite{saurav2026unistainnet}.
These methods provide effective cross-stain supervision, but their
dependence on registered or approximately corresponding images limits
the scale and diversity of available training data.

\subsection{Semi-supervised Learning for Medical Image Translation}

Semi-supervised learning combines a small collection of annotated data
with a larger set of unannotated samples and has been widely studied in
medical image classification, segmentation, and domain adaptation
\cite{cheplygina2019not,zhou2023semi,berthelot2019mixmatch}.
Representative strategies include consistency regularization
\cite{laine2016temporal,tarvainen2017mean},
pseudo-labeling \cite{lee2013pseudo}, and contrastive
representation learning \cite{chen2020simple}.

Most semi-supervised methods are designed for tasks with constrained
output spaces. In classification and segmentation, predictions on
unannotated samples can be converted into categorical or pixel-wise
supervision. Image translation is more difficult because the model must
synthesize a high-dimensional output for which multiple appearances may
seem plausible. An inaccurate generated image can therefore be
reinforced during optimization and gradually shift the learned mapping
away from the intended target domain.

This problem is particularly important in virtual staining, where the
output must simultaneously exhibit preserve the spatial organization of the source
tissue and realistic target-stain characteristics. For an unpaired source image, neither cross-stain morphology nor target-stain appearance can be directly verified using a corresponding target. Semi-supervised virtual staining therefore requires
task-specific supervision that can be obtained without additional paired images.

\subsection{Morphology Preservation and Histopathological Realism}

Morphology preservation is fundamental in virtual staining, as tissue architecture and cellular boundaries must remain spatially consistent across modalities \cite{isola2017image,wang2018high}. For paired data, reconstruction and perceptual objectives provide direct structural supervision \cite{johnson2016perceptual}. However, unpaired source images lack corresponding targets, complicating morphological fidelity in semi-supervised learning \cite{cheplygina2019not,zhou2023semi}. To address this, our framework uses Hessian-derived morphology preservation to extract structural constraints directly from the unpaired source images.

Yet, relying solely on source-derived morphological constraints can trigger optimization shortcuts \cite{geirhos2020shortcut}. Without explicit target-appearance guidance, the model may default to superficial color transformations or hallucinate contour-dominated features \cite{cohen2018distribution}. Thus, histopathological realism constraints are necessary to ensure plausible target staining. 
Although pathology foundation models capture rich diagnostic concepts \cite{hu2024boosting,saurav2026unistainnet,huang2023plip}, their use as fixed references for target-stain realism in semi-supervised translation remains largely underexplored.

Our dual-pathway approach resolves these complementary needs. Hessian-derived preservation ensures source-side spatial consistency, while realism constraints provide target-side semantic guidance. 
This interaction prevents the structural supervision from degrading into simple color-mapping. 
Ultimately, this synergy enables stable semi-supervised virtual staining that preserves both source morphology and target diagnostic semantics.
The methodology follows below.

\section{Methodology}
\label{sec:method}

\subsection{Problem Formulation}

Let $S$ and $T$ denote the source- and target-staining domains,
respectively. Virtual staining aims to learn a generator called
$G:S\rightarrow T$ that maps a source-domain image to its corresponding
target-stained image.

Then, we consider a paired dataset composed of registered source--target
images:
\begin{equation}
D_p=\left\{(s_i,t_i)\right\}_{i=1}^{N_p},
\end{equation}
where $s_i\in S$ and $t_i\in T$. Given a paired source image, the
generator produces:
\begin{equation}
\hat{t}_i=G(s_i),
\end{equation}
which can be directly supervised by the corresponding target $t_i$.
Such paired samples provide cross-stain supervision for both tissue
correspondence and target-specific staining characteristics.

Additionally, we use a collection of unpaired source images:
\begin{equation}
D_u=\left\{u_j\right\}_{j=1}^{N_u},
\end{equation}
where $u_j\in S$ has no corresponding target-stain image. Its generated
result is denoted by:
\begin{equation}
\hat{t}_j^{u}=G(u_j).
\end{equation}

Here, \emph{unpaired} specifically refers to source-domain images
without corresponding target images. Our objective is to learn the
cross-stain mapping from $D_p$ while deriving reliable supervision for
$D_u$ from the source morphology and the expected target-stain
appearance.

\begin{figure*}[t]
    \centering
    \includegraphics[width=1.0\linewidth]{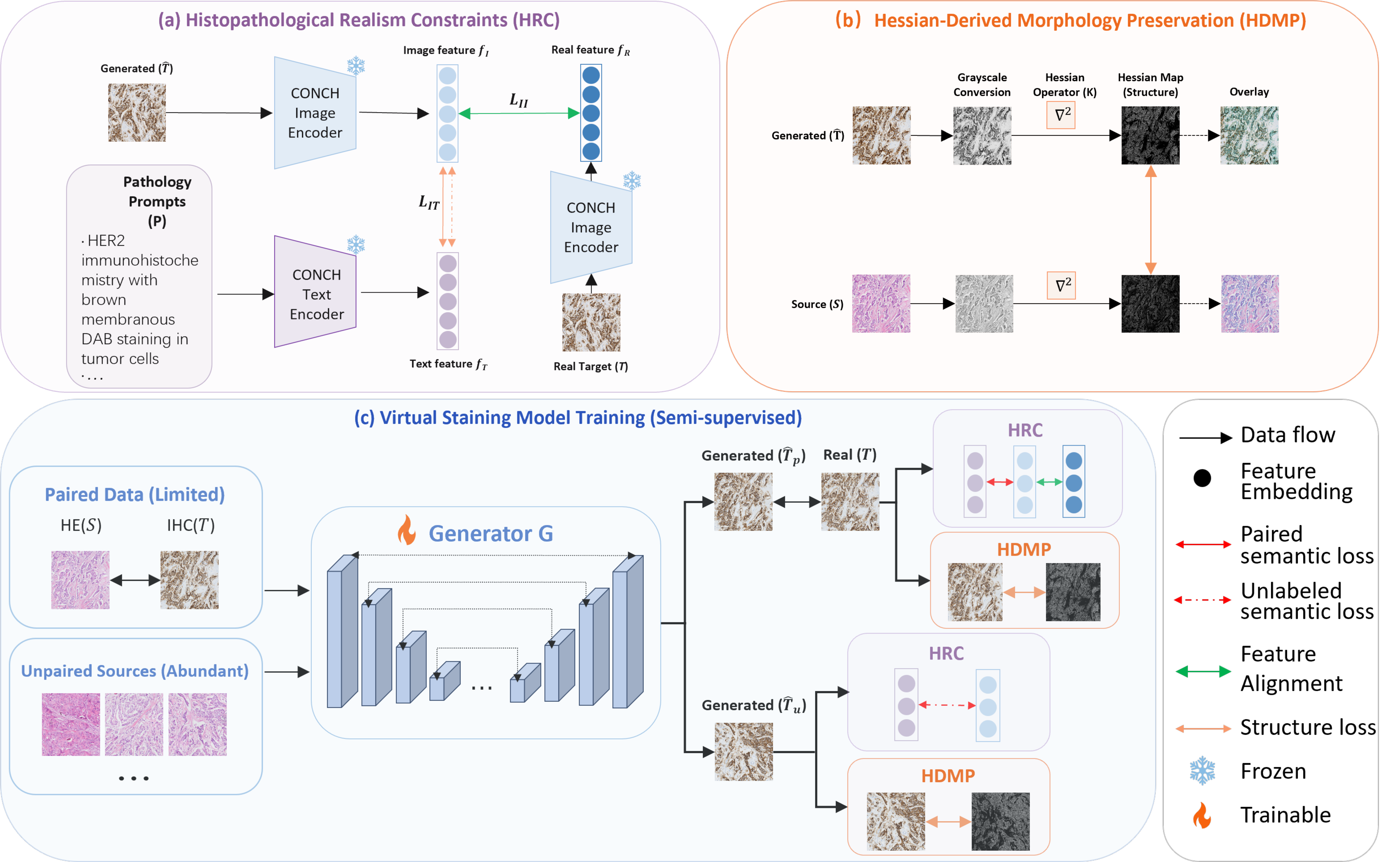}
    \caption{
    Overview of the proposed semi-supervised virtual staining
    framework.
    (a) Hessian-derived morphology preservation (HDMP) extracts structural
    supervision from each source image and constrains the generated
    output to retain source-tissue morphology.
    (b) Histopathological realism constraints (HRC) use a frozen CONCH model
    to guide the output toward plausible target-stain characteristics,
    with additional target-image alignment when a paired target is
    available.
    (c) The two supervisory pathways jointly constrain the shared
    generator. Paired samples additionally receive direct cross-stain
    supervision, whereas unpaired samples are optimized without
    corresponding target images.
    }
    \label{fig:overview_semi}
\end{figure*}

\subsection{Framework Overview}

As illustrated in Fig.~\ref{fig:overview_semi}, the proposed framework
contains paired and unpaired branches that share the same generator.
Their difference lies in the available sources of supervision.

For a paired sample $(s_i,t_i)$, the corresponding target $t_i$
provides direct cross-stain supervision for the generated image
$\hat{t}_i$. The paired branch additionally employs Hessian-derived
morphology preservation (HDMP) and histopathological realism constraints (HRC) to reinforce source-tissue consistency and target-stain
realism.

For an unpaired source image $u_j$, neither pixel-level translation
supervision nor alignment with a corresponding target image is
available. We therefore construct two complementary supervisory
pathways. HDMP extracts structural
information directly from $u_j$, providing source-side supervision
without requiring a target image. HRC supplies a target-side reference
through the frozen CONCH model and encourages $\hat{t}_j^{u}$ to exhibit
plausible target-stain characteristics.

The two methods are jointly required. HDMP specifies
which tissue structures should remain spatially consistent, but by
itself may overemphasize contours or encourage a simple color
transformation. HRC specifies what the generated target staining should
look like and discourages such structure-only shortcuts. Their
interaction provides complementary source-side and target-side
supervision for stable learning from unpaired images.

\subsection{Hessian-Derived Morphology Preservation}

An unpaired source image provides no corresponding target from which
structural consistency can be directly evaluated. Nevertheless, the
source image itself contains tissue architecture and cellular
organization that should remain spatially consistent after stain
translation. We therefore derive morphology supervision directly from
each source image.

Rapid local intensity changes in pathological images commonly occur
around tissue interfaces, glandular boundaries, cellular contours, and
nuclear edges. These structures can be characterized using second-order
spatial responses and compared between source and generated images
despite their different staining appearances.

For an image $I$, we first obtain its grayscale representation through
a fixed transformation $\mathcal{G}(\cdot)$. Its Hessian-derived
response is defined as:
\begin{equation}
R(I)
=
\operatorname{tr}
\left(
\nabla^2\mathcal{G}(I)
\right),
\end{equation}
where $\nabla^2$ denotes the spatial Hessian and
$\operatorname{tr}(\cdot)$ denotes its trace. Thereafter, for a two-dimensional
image $I$:
\begin{equation}
R(I)
=
\frac{\partial^2 \mathcal{G}(I)}{\partial x^2}
+
\frac{\partial^2 \mathcal{G}(I)}{\partial y^2}.
\end{equation}

\begin{figure}[t]
    \centering
    \includegraphics[width=\linewidth]
    {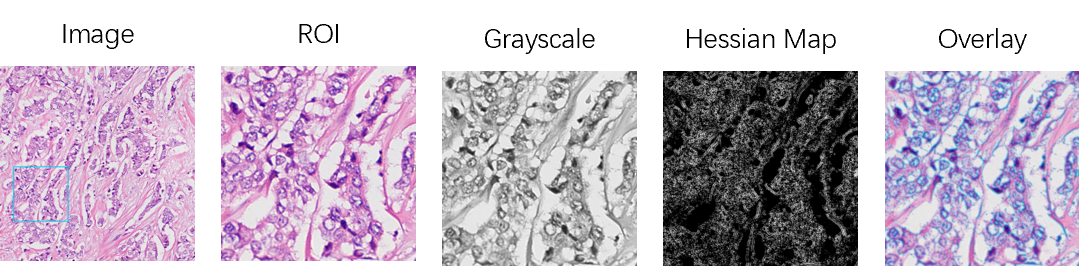}
    \caption{
    Visualization of the Hessian-derived morphology map.
    A representative pathological region is converted to grayscale
    and processed using the fixed Hessian-trace operator. Strong
    responses are mainly observed around tissue interfaces, cellular
    contours, and nuclear boundaries. The overlay is shown only for
    visualization.
    }
    \label{fig:hessian_response}
\end{figure}

As shown in Fig.~\ref{fig:hessian_response}, grayscale conversion
reduces direct dependence on individual color channels, while the
Hessian-derived response emphasizes morphology-related structures
around tissue interfaces, cellular contours, and nuclear boundaries.
The overlay is shown only for visualization, whereas the loss is
computed from the original signed response maps.

Morphology preservation compares the response of each generated image
with that of its original source image. 
Hence, for a paired sample:
\begin{equation}
L_{\mathrm{mor}}^{p}
=
\left\|
R(\hat{t}_i)-R(s_i)
\right\|_1.
\end{equation}

On the contrary, for an unpaired source image:
\begin{equation}
L_{\mathrm{mor}}^{u}
=
\left\|
R(\hat{t}_j^{u})-R(u_j)
\right\|_1.
\end{equation}

Unlike direct RGB reconstruction, this method does not require the
source and generated images to share the same staining appearance.
Instead, it transfers morphology-related supervision from the source
image to the generated result, reducing changes to tissue interfaces,
cellular boundaries, and nuclear contours.

Morphology preservation alone, however, does not determine whether the
generated image exhibits realistic target staining. Excessive reliance
on the source-derived response may produce contour-dominated outputs or
reduce translation to a superficial color transformation. A
complementary target-side constraint is therefore required.

\subsection{Histopathological Realism Constraints}

To complement the source-derived morphology supervision, we introduce
histopathological realism constraints based on the frozen CONCH
vision--language model. HRC guides generated images toward plausible
target-stain appearances and modality-specific staining
characteristics.

A task-specific textual prompt $P$ describes the expected appearance
and characteristic staining pattern of the target modality. Its
normalized text representation is

\begin{equation}
z_T=
\frac{E_T(P)}
{\left\|E_T(P)\right\|_2},
\end{equation}

where $E_T(\cdot)$ denotes the frozen CONCH text encoder. Given a
generated image $\hat{t}$, its normalized visual representation is given by:
\begin{equation}
z_I(\hat{t})=
\frac{E_I(\hat{t})}
{\left\|E_I(\hat{t})\right\|_2},
\end{equation}
where $E_I(\cdot)$ denotes the frozen CONCH image encoder.

The image--text realism constraint is defined as:
\begin{equation}
L_{\mathrm{IT}}(\hat{t},P)
=
\log
\left[
1+
\exp
\left(
-\frac{
z_I(\hat{t})^{\top}z_T
}{\tau}
\right)
\right],
\end{equation}
where $\tau$ is a temperature parameter. This objective encourages the
generated representation to remain compatible with the expected
target-stain characteristics.

\subsubsection{Paired Realism Constraint}

For a paired sample $(s_i,t_i)$, the textual reference is supplemented
with sample-specific guidance from the corresponding target image. Its
normalized representation is:
\begin{equation}
z_R(t_i)=
\frac{E_I(t_i)}
{\left\|E_I(t_i)\right\|_2}.
\end{equation}

The image--image alignment constraint is defined as:
\begin{equation}
L_{\mathrm{II}}^{p}
=
\left\|
z_I(\hat{t}_i)-z_R(t_i)
\right\|_2^2.
\end{equation}

The complete HRC objective for the paired branch is:
\begin{equation}
L_{\mathrm{real}}^{p}
=
L_{\mathrm{IT}}(\hat{t}_i,P)
+
\lambda_f L_{\mathrm{II}}^{p}.
\end{equation}

The image--text term provides a general reference for the expected
target-stain appearance, whereas image--image alignment supplies
sample-specific cross-stain information from the corresponding real
target.

\subsubsection{Unpaired Realism Constraint}

For an unpaired source image $u_j$, no corresponding target is
available for image--image alignment. HRC therefore relies on the
target-stain prompt:
\begin{equation}
L_{\mathrm{real}}^{u}
=
L_{\mathrm{IT}}(\hat{t}_j^{u},P).
\end{equation}

Because both CONCH encoders remain frozen, the target-stain reference
does not change during generator optimization. This stable reference
reduces appearance drift and complements the morphology supervision
derived from the source image. Together, they discourage both
structurally distorted results and contour-preserving but unrealistic
stain transformations.

\subsection{Training Objective}

For a paired sample, the conventional image-to-image translation
objective is computed by:
\begin{equation}
L_{\mathrm{trans}}
=
L_{\mathrm{GAN}}
+
\lambda_{\mathrm{rec}}L_{\mathrm{rec}}.
\end{equation}

The complete paired-branch objective combines direct cross-stain
supervision, morphology preservation, and histopathological realism
constraints by:
\begin{equation}
L_p
=
L_{\mathrm{trans}}
+
\lambda_m L_{\mathrm{mor}}^{p}
+
\lambda_r L_{\mathrm{HRC}}^{p}.
\end{equation}

For an unpaired source image, no direct translation loss or
target-image alignment can be computed. Its objective becomes therefore:
\begin{equation}
L_u
=
\lambda_m L_{\mathrm{mor}}^{u}
+
\lambda_r L_{\mathrm{real}}^{u}.
\end{equation}

The mini-batch objective is selected according to the availability of
corresponding target images:
\begin{equation}
L
=
\begin{cases}
L_p, & \text{for a paired batch},\\[2mm]
L_u, & \text{for an unpaired batch}.
\end{cases}
\end{equation}

In practice, the generator is first warmed up using paired samples,
after which the unpaired branch is activated. This schedule prevents
unpaired optimization from being introduced before a basic cross-stain
mapping has been established. The activation point and loss-weight
settings are provided in the Sec. Implementation Details. The proposed
methods introduce complementary training signals without modifying the
generator architecture.

\section{Experiments}
\label{sec:experiments}

\subsection{Datasets}

We evaluate the proposed framework on three virtual staining tasks:
H\&E-to-IHC virtual staining for Ki67 and HER2, and FFPE-to-H\&E
translation. For each task, accurately registered source--target pairs
provide paired supervision, while additional unpaired source images are
incorporated during semi-supervised training.

\subsubsection{H\&E-to-IHC Virtual Staining for Ki67}

The paired data are selected from the public MIST-Ki67 dataset. To
ensure reliable supervision, we retain only image pairs with satisfactory
registration quality, resulting in 2,000 pairs for training and 1,000
pairs for testing. In addition, 10,000 source-domain patches from the
Ki67 subset of the public IHC4BC dataset are used as unpaired source
images.

\subsubsection{H\&E-to-IHC Virtual Staining for HER2}

We use a private paired dataset, self-HER2, constructed from 30
H\&E--HER2 IHC WSI pairs stained with the 4B5 antibody and covering
HER2 scores of 0, 1+, 2+, and 3+. Following WSI registration using
Gatenbee et al.'s method \cite{gatenbee2023virtual}, patch extraction
at $1024\times1024$ pixels, and manual quality control, 3,000
registered pairs are retained, including 2,500 for training and 500
for testing. Additionally, 10,000 source-domain patches from the HER2
subset of IHC4BC are used as unpaired source images.

\subsubsection{FFPE-to-H\&E Translation}

For FFPE-to-H\&E translation, we use a private paired dataset,
self-FFPE. Following WSI registration and manual quality control,
2,500 registered pairs are retained for training and 1,000 pairs for
testing. A separate private dataset, self-FFPE2, provides 7,500
unpaired FFPE patches for semi-supervised training.

\subsection{Implementation Details}

Unless otherwise specified, the unpaired-to-paired data ratio is set
to 2:1, and all models are trained for 100 epochs. The generator is
first warmed up using paired samples for 70 epochs, after which
unpaired samples are introduced for the remaining 30 epochs. For
Hessian-derived morphology preservation, RGB images are converted to
grayscale using $0.299R+0.587G+0.114B$ and processed with a fixed
four-neighbor Laplacian kernel
$\left[\begin{smallmatrix}
0&1&0\\
1&-4&1\\
0&1&0
\end{smallmatrix}\right]$.
The loss is computed from the raw signed responses, while display
enhancement is used only for visualization. For comparison with
Pix2PixHD, the generator backbone, optimizer, and total number of
training epochs are kept identical. Other methods follow their official
implementations and use the same data splits. All experiments are
conducted on eight NVIDIA RTX 3090 GPUs with 24 GB memory each.

\subsection{Comparison Methods}

We compare the proposed method with representative virtual staining
approaches, including Pix2Pix, Pix2PixHD, ASP, PyramidPix2Pix,
TDKStain, HistDiST, and UNIStainNet. These methods cover supervised
image-to-image translation, weakly paired learning, diffusion-based
generation, and pathology foundation model-based approaches.

We additionally combine the proposed framework with Pix2PixHD to
evaluate its effect under the same generator backbone. TDKStain and
HistDiST are specifically designed for H\&E-to-IHC translation and
are therefore evaluated only on the H\&E-to-IHC datasets.

\subsection{Evaluation Protocol}

We evaluate the generated images from three complementary perspectives:
image-level quality, stain-specific agreement, and downstream diagnostic
utility.

\subsubsection{Image-Level Quality}

PSNR, SSIM, and MS-SSIM measure pixel-level and structural agreement
between generated and reference images. DISTS evaluates perceptual
similarity, while FID and KID measure distributional agreement.

\subsubsection{Stain-Specific Assessment}

For H\&E-to-IHC virtual staining, DAB-KL and integrated optical
density difference (IOD-D) are used to assess biomarker-related
staining. Generated and real IHC images are decomposed to obtain their
DAB channels. DAB-KL measures the divergence between their DAB
distributions, whereas IOD-D measures the difference in total
DAB-positive staining intensity. Lower values indicate closer agreement
with the real IHC images.

For FFPE-to-H\&E translation, generated and real images are transformed
into the HED color space. Dice scores computed from the hematoxylin and
eosin channels, denoted as H-Dice and E-Dice, evaluate the agreement of
nuclei-related and eosin-stained tissue regions, respectively.

\subsubsection{Downstream Diagnostic Evaluation}

For H\&E-to-IHC virtual staining, we further examine whether generated
IHC images preserve information required for four-class HER2 scoring.
The evaluation is conducted on an external dataset with HER2
annotations. Original H\&E images and real IHC images are included as
lower- and upper-reference inputs, respectively.

\subsection{Main Results}

\subsubsection{Comparison with Baselines and SOTA Methods}

The quantitative results are reported in
Tables~\ref{tab:her2-ki67} and~\ref{tab:ffpe}. Across H\&E-to-IHC
virtual staining for HER2 and Ki67, as well as FFPE-to-H\&E
translation, incorporating the proposed framework consistently
improves Pix2PixHD across image-level, distributional, and
stain-specific evaluations.

Compared with representative virtual staining methods, our framework
achieves competitive or superior performance across the three tasks,
with improvements extending beyond visual similarity to biomarker-related
staining and tissue-component agreement.

\begin{figure*}[t]
    \centering
    \includegraphics[width=0.95\linewidth]{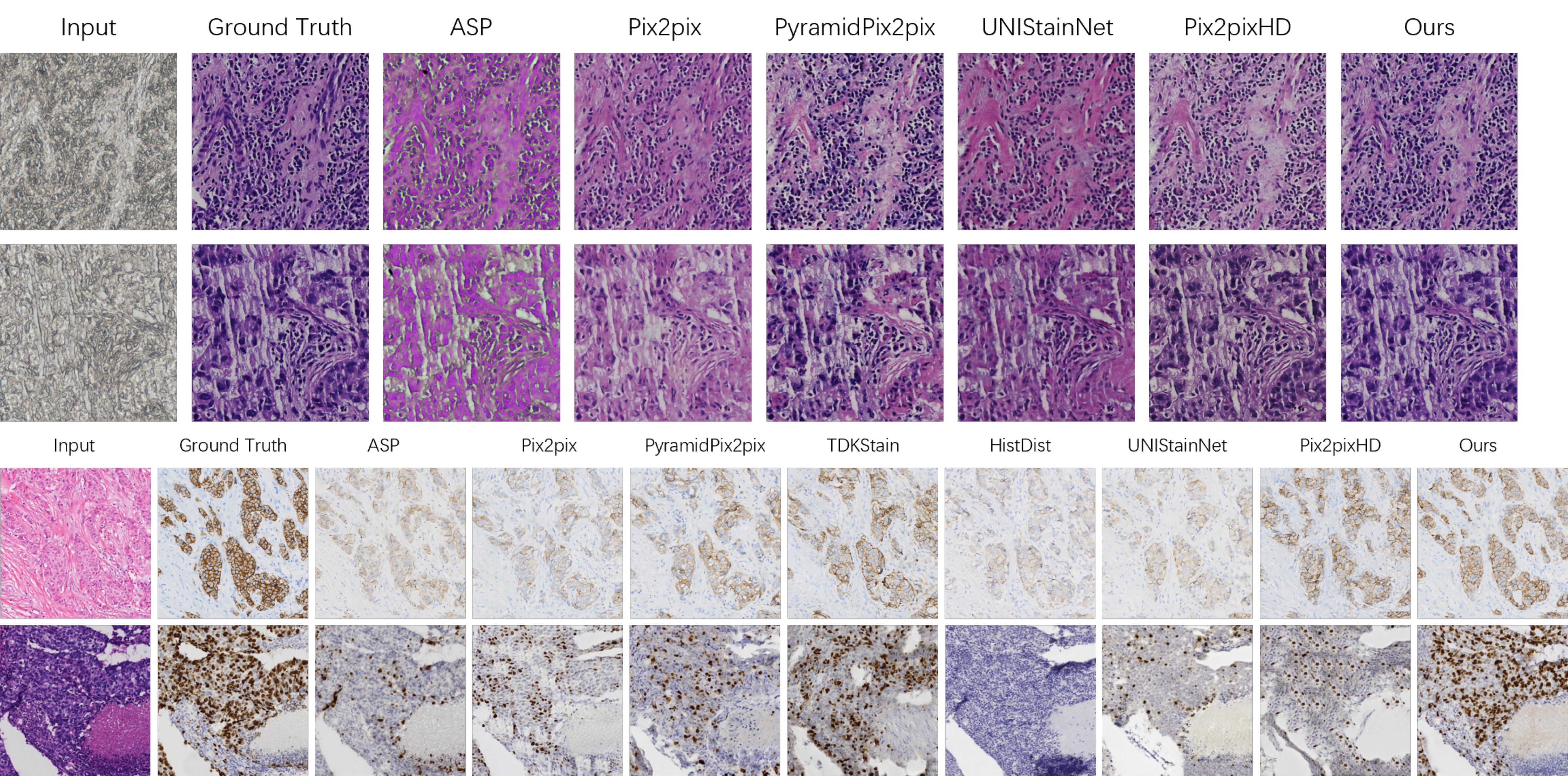}
    \caption{
Representative qualitative results on the three virtual staining
tasks. The first two rows show FFPE-to-H\&E translation, while the
third and fourth rows show H\&E-to-IHC virtual staining for HER2 and
Ki67, respectively.
}
    \label{fig:vis}
\end{figure*}

Representative results are shown in Fig.~\ref{fig:vis}. Our method
produces more plausible target-stain appearances while better retaining
tissue organization and cellular structures. These results reveal
the complementary roles of the two proposed methods: Hessian-derived
morphology preservation transfers structural supervision from the
source image, while histopathological realism constraints prevent the
translation from degenerating into contour enhancement or superficial
color transformation.

\subsubsection{Downstream Diagnostic Evaluation}

To evaluate whether the generated IHC images preserve diagnostically
relevant information, we conduct a downstream four-class HER2 scoring
experiment on an external dataset with explicit HER2 annotations.
Original H\&E images and real IHC images are included as lower- and
upper-reference inputs, respectively, while the remaining settings use
virtual IHC images generated by different methods.

\begin{figure}[h]
    \centering
    \includegraphics[width=\linewidth]{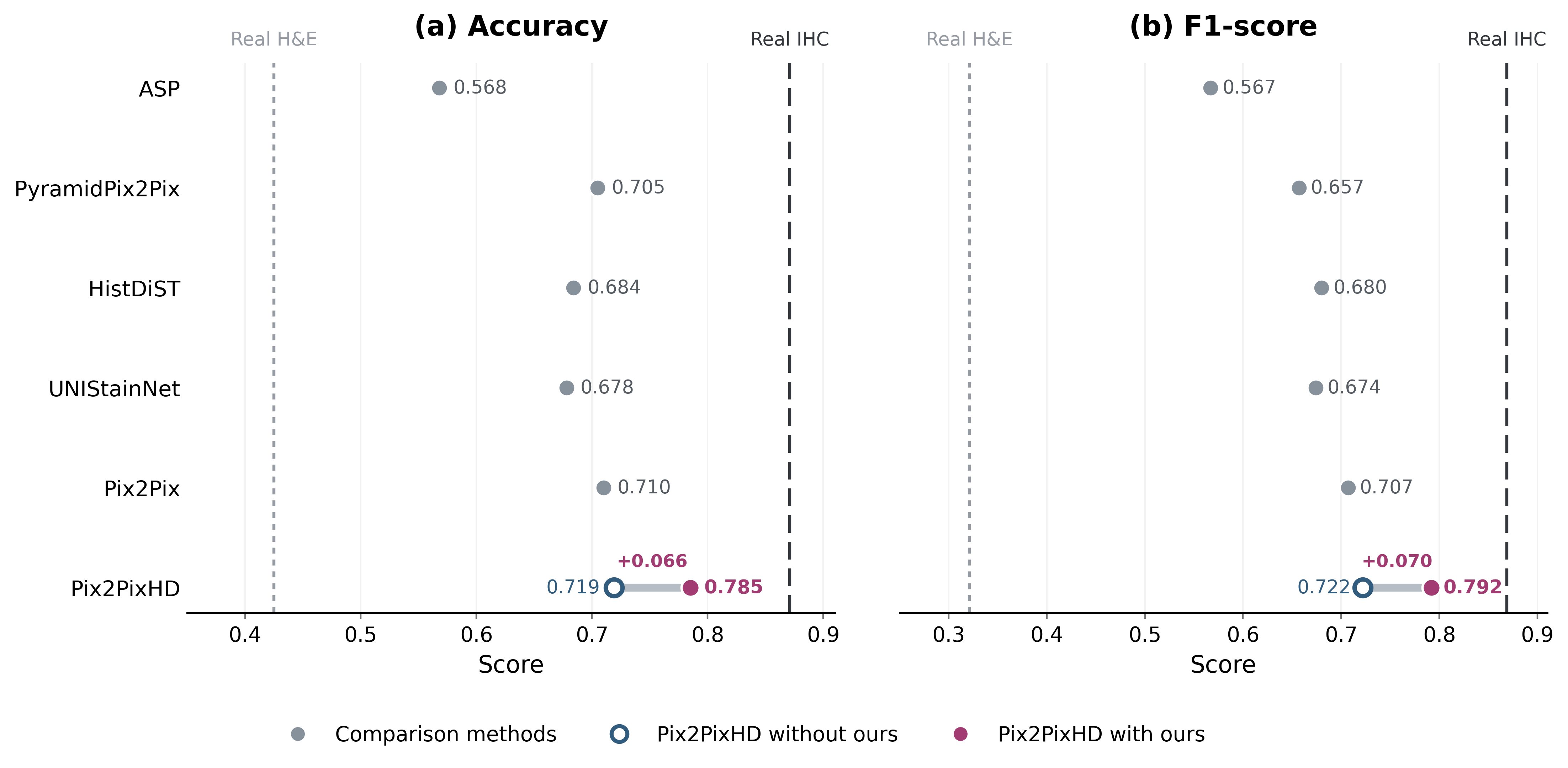}
    \caption{
    Four-class HER2 scoring performance using generated IHC images. Original H\&E images and real IHC images are included as lower- and upper-reference inputs, respectively.
    }
    \label{fig:downstream1}
\end{figure}

As shown in Fig.~\ref{fig:downstream1}, the proposed method achieves the
strongest overall HER2 scoring performance among the compared virtual
staining methods. This result indicates that the generated images exhibit realistic
staining appearances while preserving HER2-related information useful
for downstream diagnostic assessment.

\begin{table*}[htbp]
  \centering
  \caption{
Quantitative comparison on H\&E-to-IHC virtual staining for HER2
and Ki67 using the self-HER2 and MIST-Ki67 test sets.
}
    \begin{adjustbox}{width=0.95\textwidth}   
    \begin{tabular}{llcccccccc}
    \toprule
          &       & \multicolumn{3}{c}{Image-Level Agreement} & \multicolumn{3}{c}{Perceptual / Distributional Quality} & \multicolumn{2}{c}{Stain-Specific Agreement} \\
           \cmidrule(r){3-5} \cmidrule(r){6-8} \cmidrule(r){9-10} 
          &       & PSNR$\uparrow$  & SSIM$\uparrow$  & MSSSIM$\uparrow$ & FID$\downarrow$   & KID$\downarrow$   & DISTS$\downarrow$ & DAB-KL$\downarrow$ & IOD-D$\downarrow$ \\
    
    \midrule
    \multirow{8}[4]{*}{Self-HER2} & ASP   & 16.38 & 0.2304 & 0.2735 & 155.6165 & 0.1459 & 0.2847 & 0.5660 & 0.0263 \\
          & PyramidPix2Pix & 17.95 & 0.3387 & 0.3746 & 90.1506 & 0.0680 & 0.2545 & 0.3868 & 0.0243 \\
          & TDKStain & 17.89 & 0.3465 & 0.3653 & 64.7646 & 0.0298 & 0.2954 & 0.3812 & 0.0255 \\
          & HistDiST & 18.14 & 0.3345 & 0.3694 & 56.0305 & 0.0245 & 0.2542 & 0.4599 & 0.0269 \\
          & UNIStainNet & 17.82 & 0.3528 & 0.3870 & 59.4592 & 0.0259 & 0.2418 & 0.4180 & 0.0256 \\
          & Pix2Pix & 17.92 & 0.3487 & 0.3854 & 79.8819 & 0.0503 & 0.2423 & 0.4116 & 0.0246 \\
          & Pix2PixHD & 18.39 & 0.3321 & 0.3726 & 87.6770 & 0.0660 & 0.2807 & 0.4773 & 0.0251 \\
\cmidrule{2-10}          & Pix2PixHD w/ ours & 18.60 & 0.3704 & 0.4029 & 52.4523 & 0.0249 & 0.2335 & 0.3659 & 0.0240 \\
    \midrule
    \multirow{8}[4]{*}{MIST-Ki67} & ASP   & 14.60 & 0.2299 & 0.1836 & 51.1829 & 0.0295 & 0.2438 & 0.5622 & 0.0363 \\
          & PyramidPix2Pix & 14.23 & 0.2287 & 0.1698 & 91.3043 & 0.0790 & 0.2611 & 0.5113 & 0.0331 \\
          & TDKStain & 14.47 & 0.2388 & 0.1802 & 55.7503 & 0.0347 & 0.2374 & 0.4990 & 0.0313 \\
          & HistDiST & 14.41 & 0.2355 & 0.1868 & 77.8673 & 0.0480 & 0.2818 & 0.8274 & 0.0425 \\
          & UNIStainNet & 14.50 & 0.1858 & 0.1741 & 54.5161 & 0.0323 & 0.3020 & 0.5490 & 0.0389 \\
          & Pix2Pix & 13.79 & 0.2462 & 0.1827 & 78.0272 & 0.0538 & 0.2634 & 0.5969 & 0.0430 \\
          & Pix2PixHD & 14.21 & 0.2356 & 0.1727 & 114.7570 & 0.0936 & 0.2871 & 0.5827 & 0.0377 \\
\cmidrule{2-10}          & Pix2PixHD w/ ours & 14.90 & 0.2651 & 0.2122 & 47.2794 & 0.0277 & 0.2595 & 0.4969 & 0.0305 \\
    \bottomrule
    \end{tabular}%
    \end{adjustbox}
  \label{tab:her2-ki67}%
\end{table*}

\begin{table*}[ht]
  \centering
  \caption{
Quantitative comparison on FFPE-to-H\&E translation using the
self-FFPE test set.
}
    \begin{adjustbox}{width=0.95\textwidth} 
    \begin{tabular}{llcccccccc}
    \toprule
          &       & \multicolumn{3}{c}{Image-Level Agreement} & \multicolumn{3}{c}{Perceptual / Distributional Quality} & \multicolumn{2}{c}{Stain-Specific Agreement} \\
          \cmidrule(r){3-5} \cmidrule(r){6-8} \cmidrule(r){9-10} 
          &       & PSNR$\uparrow$  & SSIM$\uparrow$  & MSSSIM$\uparrow$ & FID$\downarrow$   & KID$\downarrow$   & DISTS$\downarrow$ & H-Dice$\uparrow$ & E-Dice$\uparrow$ \\
    \midrule
    \multirow{6}[4]{*}{Self-FFPE} & ASP   & 15.79 & 0.3583 & 0.3632 & 97.6618 & 0.0640 & 0.2895 & 0.1984 & 0.7604 \\
          & PyramidPix2Pix & 17.59 & 0.4852 & 0.6577 & 23.1832 & 0.0043 & 0.1422 & 0.5454 & 0.6749 \\
          & UNIStainNet & 18.15 & 0.5228 & 0.6892 & 19.5909 & 0.0017 & 0.2267 & 0.5893 & 0.7387 \\
          & Pix2Pix & 15.14 & 0.3736 & 0.4165 & 28.3412 & 0.0053 & 0.1794 & 0.3217 & 0.4537 \\
          & Pix2PixHD & 17.27 & 0.5034 & 0.6688 & 21.2485 & 0.0032 & 0.1164 & 0.5168 & 0.6241 \\
\cmidrule{2-10}
          & Pix2PixHD w/ ours & 18.45 & 0.5377 & 0.7028 & 16.0702 & 0.0005 & 0.1123 & 0.6414 & 0.7730 \\
    \bottomrule
    \end{tabular}%
    \end{adjustbox}
  \label{tab:ffpe}%
\end{table*}%

\begin{table*}[htbp]
  \centering
  \caption{
Robustness comparison on the self-HER2 and MIST-Ki67 test sets under
staining variations and acquisition degradations.
}
   \begin{adjustbox}{width=0.95\textwidth} 
    \begin{tabular}{llcccccccccccccc}
    \toprule
          &       & \multicolumn{7}{c}{Staining Variation}  & \multicolumn{7}{c}{Acquisition Degradation} \\
          \cmidrule(r){3-9} \cmidrule(r){10-16}
          &       & PSNR$\uparrow$  & SSIM$\uparrow$  & FID$\downarrow$ & KID$\downarrow$ & DISTS$\downarrow$ & DAB-KL$\downarrow$ & IOD-D$\downarrow$   & PSNR$\uparrow$  & SSIM$\uparrow$  & FID$\downarrow$ & KID$\downarrow$ & DISTS$\downarrow$ & DAB-KL$\downarrow$ & IOD-D$\downarrow$ \\
    \midrule
    \multirow{7}[2]{*}{\makecell{Self-HER2}} & ASP & 15.52 & 0.2107 & 117.8376 & 0.0998 & 0.2939 & 0.7326 & 0.0361 & 16.2876 & 0.2389 & 157.5783 & 0.1403 & 0.2842 & 0.5559 & 0.0267 \\
          & PyramidPix2Pix & 17.39 & 0.3127 & 107.6362 & 0.0918 & 0.2868 & 0.4717 & 0.0270 & 17.9899 & 0.3321 & 93.2739 & 0.0722 & 0.2815 & 0.4064 & 0.0247 \\
          & TDKStain & 17.42 & 0.2860 & 80.3787 & 0.0440 & 0.2894 & 0.5423 & 0.0246 & 17.8115 & 0.3214 & 64.7279 & 0.0305 & 0.2808 & 0.3896 & 0.0258 \\
          & UNIStainNet & 17.37 & 0.2317 & 68.8050 & 0.0366 & 0.3211 & 0.4289 & 0.0252 & 17.0465 & 0.2524 & 63.4894 & 0.0316 & 0.3355 & 0.4165 & 0.0271 \\
          & Pix2Pix & 17.65 & 0.3254 & 78.6544 & 0.0519 & 0.2758 & 0.4154 & 0.0251 & 17.2000 & 0.3453 & 97.7118 & 0.0466 & 0.2541 & 0.4185 & 0.0245 \\
          & Pix2PixHD & 18.17 & 0.3114 & 88.4236 & 0.0652 & 0.2791 & 0.4667 & 0.0252 & 18.3600 & 0.3499 & 86.3794 & 0.0586 & 0.2661 & 0.4866 & 0.0253 \\
          & Pix2PixHD w/ ours & 18.40 & 0.3403 & 59.1927 & 0.0264 & 0.2613 & 0.3896 & 0.0249 & 18.4700 & 0.3619 & 59.5548 & 0.0283 & 0.2494 & 0.3821 & 0.0243 \\
    \midrule
    \multirow{7}[2]{*}{\makecell{MIST-Ki67}} & ASP & 14.23 & 0.2186 & 59.1870 & 0.0357 & 0.2608 & 0.6544 & 0.0391 & 14.5465 & 0.2330 & 59.8069 & 0.0349 & 0.2525 & 0.5929 & 0.0377 \\
          & PyramidPix2Pix & 13.97 & 0.2182 & 92.3648 & 0.0834 & 0.2644 & 0.5274 & 0.0378 & 14.3174 & 0.2321 & 90.5449 & 0.0757 & 0.2658 & 0.5181 & 0.0338 \\
          & TDKStain & 13.75 & 0.2127 & 61.2593 & 0.0423 & 0.2520 & 0.5811 & 0.0445 & 14.3700 & 0.2344 & 54.7415 & 0.0319 & 0.2383 & 0.5198 & 0.0321 \\
          & UNIStainNet & 14.46 & 0.1841 & 65.4040 & 0.0439 & 0.3041 & 0.5586 & 0.0390 & 14.2200 & 0.1890 & 62.9493 & 0.0337 & 0.3148 & 0.5531 & 0.0349 \\
          & Pix2Pix & 13.38 & 0.2357 & 85.2704 & 0.0546 & 0.2719 & 0.5852 & 0.0466 & 13.9137 & 0.2471 & 72.9535 & 0.0436 & 0.2689 & 0.5078 & 0.0428 \\
          & Pix2PixHD & 13.95 & 0.2266 & 123.0893 & 0.1070 & 0.2883 & 0.6755 & 0.0402 & 14.1927 & 0.2355 & 115.1264 & 0.0933 & 0.2860 & 0.5917 & 0.0379 \\
          & Pix2PixHD w/ ours & 14.74 & 0.2610 & 56.5030 & 0.0347 & 0.2599 & 0.5198 & 0.0345 & 14.8844 & 0.2737 & 49.2880 & 0.0289 & 0.2537 & 0.4907 & 0.0311 \\
    \bottomrule
    \end{tabular}%
    \end{adjustbox}
  \label{tab:her2_ki67_robust}%
\end{table*}

\begin{table*}[htbp]
  \centering
  \caption{
Robustness comparison on the self-FFPE test set under staining
variations and acquisition degradations.
}
  \begin{adjustbox}{width=0.95\textwidth}
    \begin{tabular}{clcccccccccccccc}
    \toprule
          &       & \multicolumn{7}{c}{Staining Variation}  & \multicolumn{7}{c}{Acquisition Degradation} \\
          \cmidrule(r){3-9} \cmidrule(r){10-16}
          &       & PSNR$\uparrow$  & SSIM$\uparrow$  & FID$\downarrow$ & KID$\downarrow$ & DISTS$\downarrow$ & H-Dice$\uparrow$ & E-Dice$\uparrow$ & PSNR$\uparrow$  & SSIM$\uparrow$  & FID$\downarrow$ & KID$\downarrow$ & DISTS$\downarrow$ & H-Dice$\uparrow$ & E-Dice$\uparrow$ \\
    \midrule
    \multirow{6}[2]{*}{\makecell{Self-FFPE2HE}} & ASP & 15.52 & 0.3474 & 96.9358 & 0.0609 & 0.2879 & 0.2171 & 0.7507 & 15.5092 & 0.3455 & 99.5627 & 0.0595 & 0.2908 & 0.1746 & 0.7175 \\
          & PyramidPix2Pix & 17.12 & 0.4679 & 27.8536 & 0.0067 & 0.1529 & 0.5189 & 0.6592 & 16.8787 & 0.4581 & 27.6058 & 0.0055 & 0.1574 & 0.4951 & 0.6220 \\
          & UNIStainNet   & 17.67 & 0.5151 & 21.2922 & 0.0025 & 0.2289 & 0.5735 & 0.7270 & 17.4587 & 0.4926 & 22.1434 & 0.0023 & 0.2429 & 0.5632 & 0.7253 \\
          & Pix2Pix & 13.75 & 0.3101 & 78.6262 & 0.0333 & 0.3039 & 0.2286 & 0.3185 & 14.7938 & 0.3512 & 32.0556 & 0.0058 & 0.1953 & 0.2956 & 0.4517 \\
          & Pix2PixHD & 16.91 & 0.4900 & 23.8326 & 0.0045 & 0.1251 & 0.4876 & 0.6065 & 16.4963 & 0.4664 & 24.1984 & 0.0043 & 0.1312 & 0.4623 & 0.5605 \\
          & Pix2PixHD w/ ours & 17.99 & 0.5191 & 18.1853 & 0.0013 & 0.1182 & 0.6112 & 0.7696 & 17.6075 & 0.4982 & 20.5926 & 0.0016 & 0.1230 & 0.6004 & 0.7400 \\
    \bottomrule
    \end{tabular}%
    \end{adjustbox}
  \label{tab:ffpe_robust}%
\end{table*}

\begin{table}[htbp]
\centering
\caption{
Ablation study of Hessian-derived morphology preservation and histopathological realism constraints on the self-HER2 dataset.}
\begin{adjustbox}{width=0.47\textwidth}
\begin{tabular}{lcccc}
\toprule
Method & PSNR$\uparrow$ & SSIM$\uparrow$ & FID$\downarrow$ & DISTS$\downarrow$ \\
\midrule
Supervised baseline & 18.38 & 0.3321 & 87.6770 & 0.2807 \\
Semi-supervised baseline & 11.90 & 0.2048 & 455.7836 & 0.4586 \\
Semi-supervised + HDMP & 18.37 & 0.3428 & 73.6938 & 0.3001 \\
Semi-supervised + HRC & 18.42 & 0.3688 & 64.5625 & 0.2433 \\
Ours & \textbf{18.60} & \textbf{0.3704} & \textbf{52.4523} & \textbf{0.2335} \\
\bottomrule
\end{tabular}
\end{adjustbox}
\label{tab:ablation}
\end{table}

\subsection{Ablation Study}

We conduct ablation experiments to evaluate the individual and combined
effects of HDMP and HRC. As reported in
Table~\ref{tab:ablation}, directly introducing unpaired source images
without either constraint substantially degrades performance,
demonstrating that unpaired optimization requires appropriate
supervision.

Adding HDMP or HRC individually substantially improves the
semi-supervised baseline, while their combination achieves the strongest
overall performance. HDMP transfers morphology-related supervision from
the source image, whereas HRC promotes plausible target-stain
characteristics. Their complementary effects support stable learning
from unpaired source images.

\begin{figure}[t]
    \centering
    \includegraphics[width=0.95\linewidth]{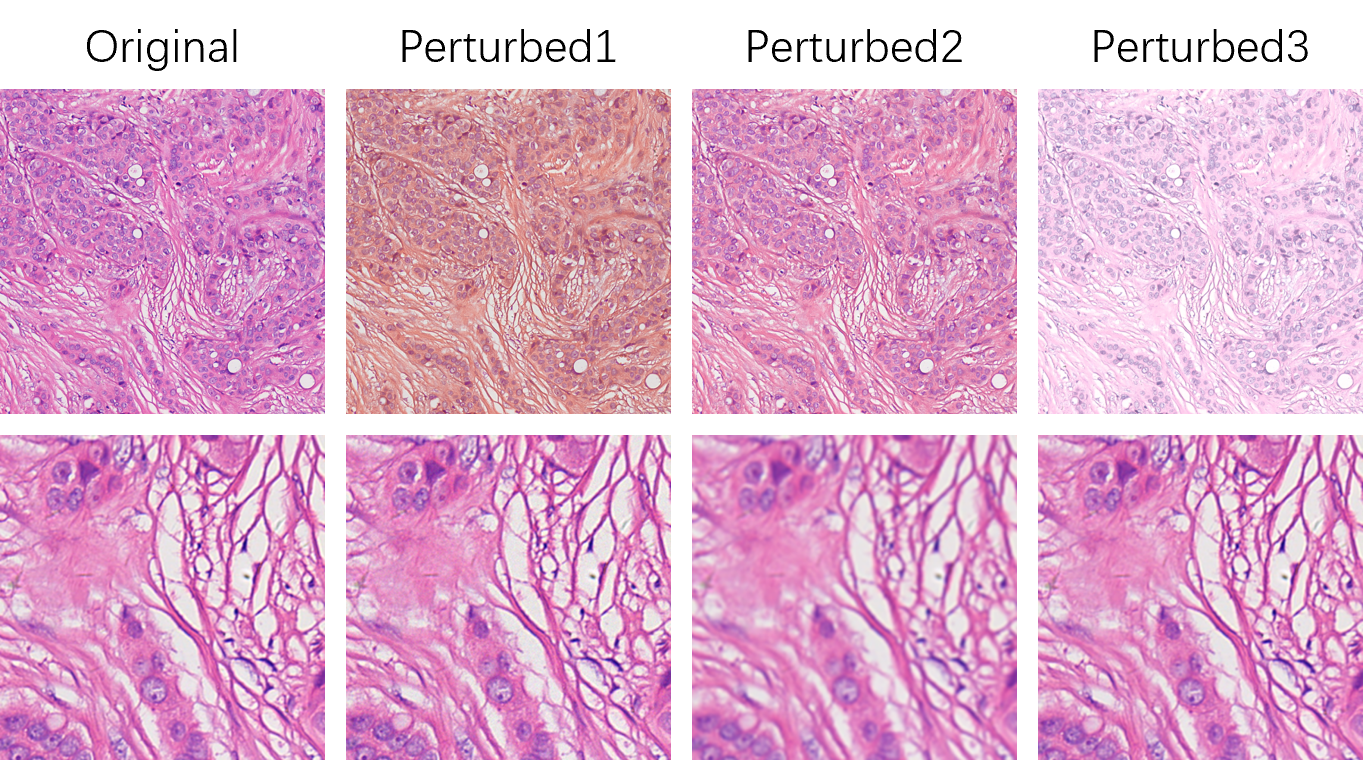}
    \caption{
Examples of input variations used for robustness evaluation.
Each row begins with the original image, followed by three modified
variants. The top and bottom rows illustrate staining variations and
acquisition degradations, respectively.
}
    \label{fig:perturbation}
\end{figure}

\begin{figure}[ht]
    \centering
    \includegraphics[width=0.95\linewidth]{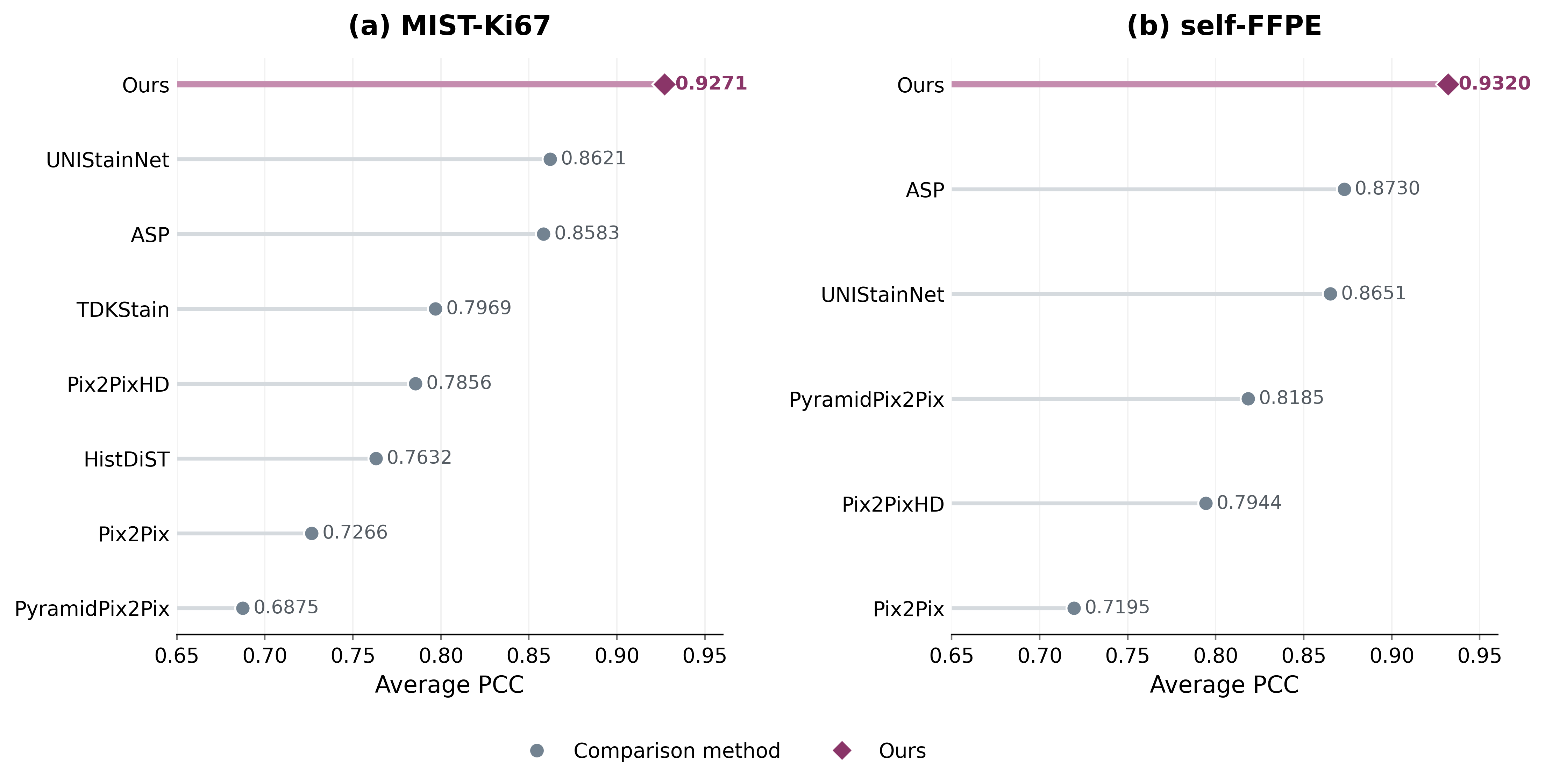}
    \caption{
    Average PCC between generated images and their corresponding source
    images on the MIST-Ki67 and self-FFPE datasets. Higher values indicate
    stronger preservation of source-tissue organization.
    }
    \label{fig:structure}
\end{figure}

\begin{figure*}[t]
    \centering
    \includegraphics[width=0.9\linewidth]{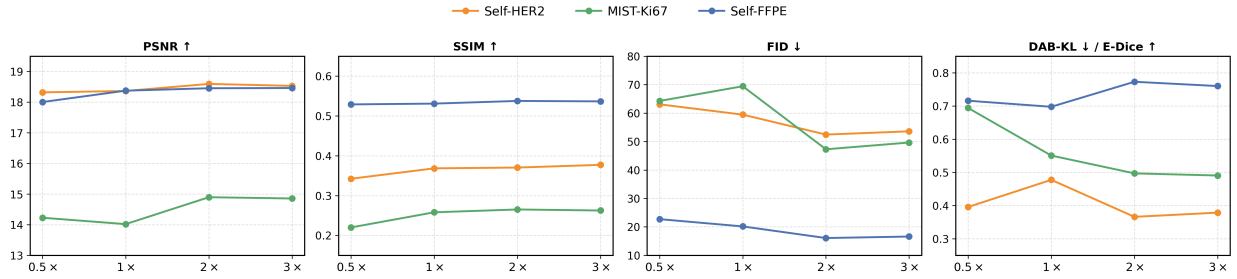}
    \caption{
    Sensitivity to the amount of unpaired source data. The
    unpaired-to-paired ratio is varied from $0.5:1$ to $3:1$ on the
    self-HER2, MIST-Ki67, and self-FFPE datasets.
    }
    \label{fig:hyperparameters}
\end{figure*}

\subsection{Robustness to Staining and Acquisition Variations}

The proposed framework emphasizes diagnostically relevant staining
characteristics and tissue morphology, which should be less sensitive
than superficial image appearance to incidental variations in staining
and image acquisition. We therefore evaluate whether the learned
translation remains stable when such variations are introduced while
the underlying pathological content is preserved.

Two modified versions of each test set are constructed to simulate
staining variations and acquisition degradations, with representative
examples shown in Fig.~\ref{fig:perturbation}. Staining variations are
generated by adjusting the hue, saturation, and brightness of the source
images. Acquisition degradations are simulated using Gaussian noise,
downsampling followed by Gaussian blur, and radial geometric distortion.
The noise standard deviation is set to 2, the downsampling factor and
blur-kernel size are set to 0.5 and 3, respectively, and the distortion
coefficient is sampled from $(-0.05,\,0.05)$. Each operation is
independently applied with a probability of 0.5.

As presented in Tables~\ref{tab:her2_ki67_robust}
and~\ref{tab:ffpe_robust}, the proposed framework consistently improves
Pix2PixHD under both types of input variation and achieves strong
overall performance across image-level, distributional, and
stain-specific evaluations. These results suggest that morphology
preservation and histopathological realism constraints encourage the
model to rely more on pathology-relevant staining and structural
information, thereby reducing its sensitivity to incidental appearance
and acquisition changes.

\subsection{Morphology Preservation Evaluation}

To evaluate whether virtual staining alters the spatial organization
of the source tissue, we compute the Pearson correlation coefficient
(PCC) between each generated image and its corresponding source image
after grayscale conversion. For H\&E-to-IHC translation, PCC is
calculated between the generated IHC image and the input H\&E image.
For FFPE-to-H\&E translation, it is calculated between the generated
H\&E image and the corresponding FFPE input.

Grayscale conversion reduces the direct influence of stain color,
while PCC measures the agreement between the spatial intensity patterns
of the two images. A higher value indicates that the generated image
retains more of the structural organization present in the source image.

As shown in Fig.~\ref{fig:structure}, the proposed framework achieves
the highest PCC on both datasets, indicating stronger preservation of
source-tissue organization. This result is consistent with the role of
HDMP in reducing changes to tissue interfaces and cellular structures
during stain translation.

\subsection{Sensitivity to the Amount of Unpaired Data}

We study the influence of unpaired data volume by setting the
unpaired-to-paired ratio to 0.5:1, 1:1, 2:1, and 3:1, while
keeping the remaining settings unchanged.

As shown in Fig.~\ref{fig:hyperparameters}, performance generally
improves as additional unpaired source images are introduced and becomes
relatively stable at a ratio of approximately 2:1. Increasing the
ratio beyond this point provides limited additional benefit. We
therefore use an unpaired-to-paired ratio of 2:1 in the remaining
experiments.

\section{Conclusion}

In this paper, we presented a stable semi-supervised framework for
virtual staining that reduces reliance on accurately registered
source--target pairs by jointly exploiting limited paired data and
abundant unpaired source images. Hessian-derived morphology preservation
extracts structural supervision from source images to preserve
morphology-related spatial structures, while histopathological realism
constraints guide generated images toward plausible target-stain
characteristics. Together, they mitigate structural and appearance drift
during semi-supervised optimization.

Experiments on H\&E-to-IHC for HER2 and Ki67, and FFPE-to-H\&E
translation demonstrate consistent improvements in image quality,
morphology preservation, pathology-informed metrics, robustness, and
downstream diagnostic performance. The robustness and ablation results
further support the stability and complementary roles of the two
proposed constraints.

Overall, the proposed framework provides a practical approach for
incorporating unpaired source images into virtual staining without
modifying the underlying generator architecture. These findings
highlight the potential of source-derived morphology supervision and
target-stain realism guidance for data-efficient and reliable virtual
staining.

\bibliographystyle{IEEEtran}
\bibliography{tmi}

\end{document}